\documentclass[10pt,twocolumn,letterpaper]{article}

\usepackage{cvpr}
\usepackage{times}
\usepackage{epsfig}
\usepackage{graphicx}        
\usepackage{amsmath}
\usepackage{amssymb}
\usepackage{color}
\usepackage{tabularx}
\usepackage{booktabs}
\usepackage{multirow}
\usepackage{algorithm}
\usepackage{algorithmicx}
\usepackage{algpseudocode}  
\usepackage{array}
\usepackage{xspace}
\usepackage{balance}
\usepackage{enumitem}
\usepackage{mathtools}
\usepackage{rotating}
\usepackage{url}
\usepackage{tabulary}
\usepackage[table]{xcolor}
\usepackage{subcaption}
\usepackage[export]{adjustbox}
\usepackage{ctable}
\usepackage{diagbox}
\usepackage{subcaption}
\newcolumntype{L}[1]{>{\raggedright\arraybackslash}p{#1}}
\newcolumntype{C}[1]{>{\centering\arraybackslash}p{#1}}
\newcolumntype{R}[1]{>{\raggedleft\arraybackslash}p{#1}}
        
\def\Vec#1{{\boldsymbol{#1}}}
\def\Mat#1{{\boldsymbol{#1}}}

\newcommand{\fig}{{Figure}\@\xspace}

\makeatletter
\newcommand{\algmargin}{\the\ALG@thistlm}   
\makeatother
\algnewcommand{\parState}[1]{\State%
	\parbox[t]{\dimexpr\linewidth-\algmargin}{\strut #1\strut}}

\graphicspath{{./figs/}}

\usepackage[pagebackref=true,breaklinks=true,letterpaper=true,colorlinks,bookmarks=false]{hyperref}

\cvprfinalcopy 

\def\cvprPaperID{2512} 

\ifcvprfinal\fi
\begin{document}

\title
	{
	Learning to Learn from Noisy Labeled Data
	}
\author{Junnan Li$^1$,~~~Yongkang Wong$^1$,~~~Qi Zhao$^2$,~~~Mohan S. Kankanhalli$^1$\\
	$^{1}$National University of Singapore~~~~~$^2$University of Minnesota\\
	{\tt\small lijunnan@u.nus.edu, yongkang.wong@nus.edu.sg, qzhao@cs.umn.edu, mohan@comp.nus.edu.sg}
}

\maketitle

\begin{abstract}
Despite the success of deep neural networks (DNNs) in image classification tasks,
the human-level performance relies on massive training data with high-quality manual annotations,
which are expensive and time-consuming to collect.
There exist many inexpensive data sources on the web,
but they tend to contain inaccurate labels.
Training on noisy labeled datasets causes performance degradation because DNNs can easily overfit to the label noise.
To overcome this problem,
we propose a noise-tolerant training algorithm,
where a meta-learning update is performed prior to conventional gradient update.
The proposed meta-learning method simulates actual training by generating synthetic noisy labels,
and train the model such that after one gradient update using each set of synthetic noisy labels,
the model does not overfit to the specific noise.
We conduct extensive experiments on the noisy CIFAR-10 dataset and the Clothing1M dataset.
The results demonstrate the advantageous performance of the proposed method compared to state-of-the-art baselines.
	
\end{abstract}

\section{Introduction}
\label{sec:introduction}

One of the key reasons why deep neural networks (DNNs) have been so successful in image classification is the collections of massive labeled datasets such as COCO \cite{coco} and ImageNet~\cite{imagenet}.
However, it is time-consuming and expensive to collect such high-quality manual annotations.
A single image often requires agreement from multiple annotators to reduce label error.
On the other hand,
there exist other less expensive sources to collect labeled data,
such as search engines, social media websites, or reducing the number of annotators per image.
However, those low-cost approaches introduce low-quality annotations with \textit{label noise}.
Many studies have shown that label noise can significantly affect the accuracy of the learned classifiers~\cite{noise_survey,Ranzato_2014_arxiv,Zhang_ICLR_2017}.
In this work, we address the following problem: how to effectively train on noisy labeled datasets?

Some methods learn with label noise by relying on human supervision to verify seed images~\cite{Lee_CVPR_2018,Andreas_CVPR_2017} or estimate label confusion~\cite{Giorgio_CVPR_2017,Tong_CVPR_2015}.
However, those methods exhibit a disadvantage in scalability for large datasets.
On the other hand, methods without human supervision (\eg label correction~\cite{Reed_2014_arxiv,Tanaka_CVPR_2018} and noise correction layers~\cite{Goldberger_ICLR_2017,Ranzato_2014_arxiv}) are scalable but less effective and more heuristic.   
In this work we propose a \textit{meta-learning} based \textit{noise-tolerant} (MLNT) training to learn from noisy labeled data without human supervision or access to any clean labels. 
Rather than designing a specific model,
we propose a model-agnostic training algorithm,
which is applicable to any model that is trained with gradient-based learning rule.

The prominent issue in training DNNs on noisy labeled data is that DNNs often overfit to the noise, which leads to performance degradation.
Our method addresses this issue by optimizing for a model's parameters that are less prone to overfitting and more robust against label noise.
Specifically, for each mini-batch,
we propose a meta-objective to train the model,
such that after the model goes through conventional gradient update, 
it does not overfit to the label noise.
The proposed meta-objective encourages the model to produce consistent predictions after it is trained on a variety of synthetic noisy labels.
The key idea of our method is:
\textit{a noise-tolerant model should be able to consistently learn the underlying knowledge from data despite different label noise}.
The main contribution of this work are as follows.
\begin{itemize}
	\item We propose a noise-tolerant training algorithm, where a meta-objective is optimized before conventional training.
	Our method can be theoretically applied to any model trained with gradient-based rule.
	We aim to optimize for a model that does not overfit to a wide spectrum of artificially generated label noise.
	\item
	We formulate our meta-objective as:
	train the model such that after it learns from various synthetic noisy labels using gradient update,
	the updated models give consistent predictions with a teacher model.	
	We adapt a self-ensembling method to construct the teacher model,
	which gives more reliable predictions unaffected by the synthetic noise.
	\item
	We perform experiments on two datasets with synthetic and real-world label noise, and demonstrate the advantageous performance of the proposed method in image classification tasks compared to state-of-the-art methods.
	In addition, we conduct extensive ablation study to examine different components of the proposed method.
	Our code is publicly available\footnote{\url{https://github.com/LiJunnan1992/MLNT}}.
\end{itemize}

\section{Related Work}
\label{sec:literature}

\noindent\textbf{Learning with label noise.}
A number of approaches have been proposed to train DNNs with noisy labeled data.
One line of approaches formulate explicit or implicit noise models to characterize the distribution of noisy and true labels,
using neural networks~\cite{Goldberger_ICLR_2017,Jiang_ICML_2018,Lee_CVPR_2018,Ren_ICML_2018,Giorgio_CVPR_2017,Ranzato_2014_arxiv,Andreas_CVPR_2017},
directed graphical models~\cite{Tong_CVPR_2015},
knowledge graphs~\cite{Li_ICCV_17},
or conditional random fields~\cite{Vahdat_NIPS_2017}.
The noise models are then used to infer the true labels or assign smaller weights to noisy samples.
However, these methods often require a small set of data with clean labels to be available,
or use expensive estimation methods.
They also rely on specific assumptions about the noise model, which may limit their effectiveness with complicated label noise.
Another line of approaches use correction methods to reduce the influence of noisy labels.  
Bootstrap method~\cite{Reed_2014_arxiv} introduces a consistency objective that effectively re-labels the data during training.
Tanaka~\etal~\cite{Tanaka_CVPR_2018} propose to jointly optimize network parameters and data labels.
An iterative training method is proposed to identify and downweight noisy samples~\cite{Wang_CVPR_2018}.
A few other methods have also been proposed that use noise-tolerant loss functions to achieve robust learning under label noise~\cite{Ghosh_AAAI_17,Ghosh_Neuro_2015,Rooyen_NIPS_15}.

\noindent\textbf{Meta-Learning.}
Recently, meta-learning methods for DNNs have resurged in its popularity.
Meta-learning generally seeks to perform the learning at a level higher than where conventional learning occurs,
\eg~learning the update rule of a learner~\cite{Ravi_Meta_2017},
or finding weight initializations that can be easily fine-tuned~\cite{Finn_MAML_2017} or transferred~\cite{Li_AAAI_2018}.
Our approach is most related to MAML~\cite{Finn_MAML_2017},
which aims to train model parameters that can learn well based on a few examples and a few gradient descent steps.
Both MAML and our method are model-agnostic and perform training by doing gradient updates on simulated meta-tasks.
However, our objective and algorithm are different from that of MAML.
MAML addresses few-shot transfer to new tasks,
whereas we aim to learn a noise-tolerant model.
Moreover, MAML trains using classification loss on a meta-test set,
whereas we use a consistency loss with a teacher model.

\noindent\textbf{Self-Ensembling.}
Several recent methods based on self-ensembling have improved the state-of-the-art results for semi-supervised learning~\cite{Laine_ICLR_2017,Sajjadi_NIPS_16,Tarvainen_NIPS_17},
where labeled samples are scarce and unlabeled samples are abundant.
These methods apply a consistency loss to the unlabeled samples,
which regularizes a neural network to make consistent predictions for the same samples under different data augmentation~\cite{Sajjadi_NIPS_16},
dropout and noise conditions~\cite{Laine_ICLR_2017}.
We focus in particular on the self-ensembling approach proposed by Tarvainen~\&~Valpola~\cite{Tarvainen_NIPS_17} as it forms one of the basis of our approach.
Their approach proposes two networks: a \textit{student} network and a \textit{teacher} network,
where the weights of the teacher are the exponential moving average of those of the student.
They enforce the student network to make consistent predictions with the teacher network.
In our method,
we use the teacher network in meta-test to train the student network such that it is more tolerant to label noise.

\section{Method}
\label{sec:method}
\begin{figure}[!t]
 \centering
  	\includegraphics[width=\columnwidth]{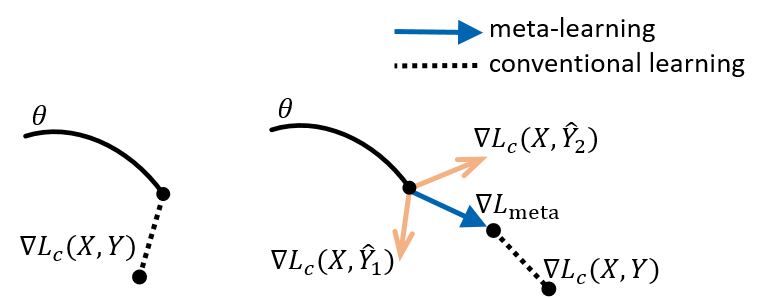}
  \caption
  {
  \small
	Left: conventional gradient update with cross entropy loss may overfit to label noise. 
	Right: a meta-learning update is performed beforehand using synthetic label noise, 
	which encourages the network parameters to be noise-tolerant and reduces overfitting during the conventional update.
  } 
  \label{fig:intuition}
 \end{figure}  	
\begin{figure*}[!t]
 \centering
  	\includegraphics[trim={0 83 0 0},clip,width=\textwidth]{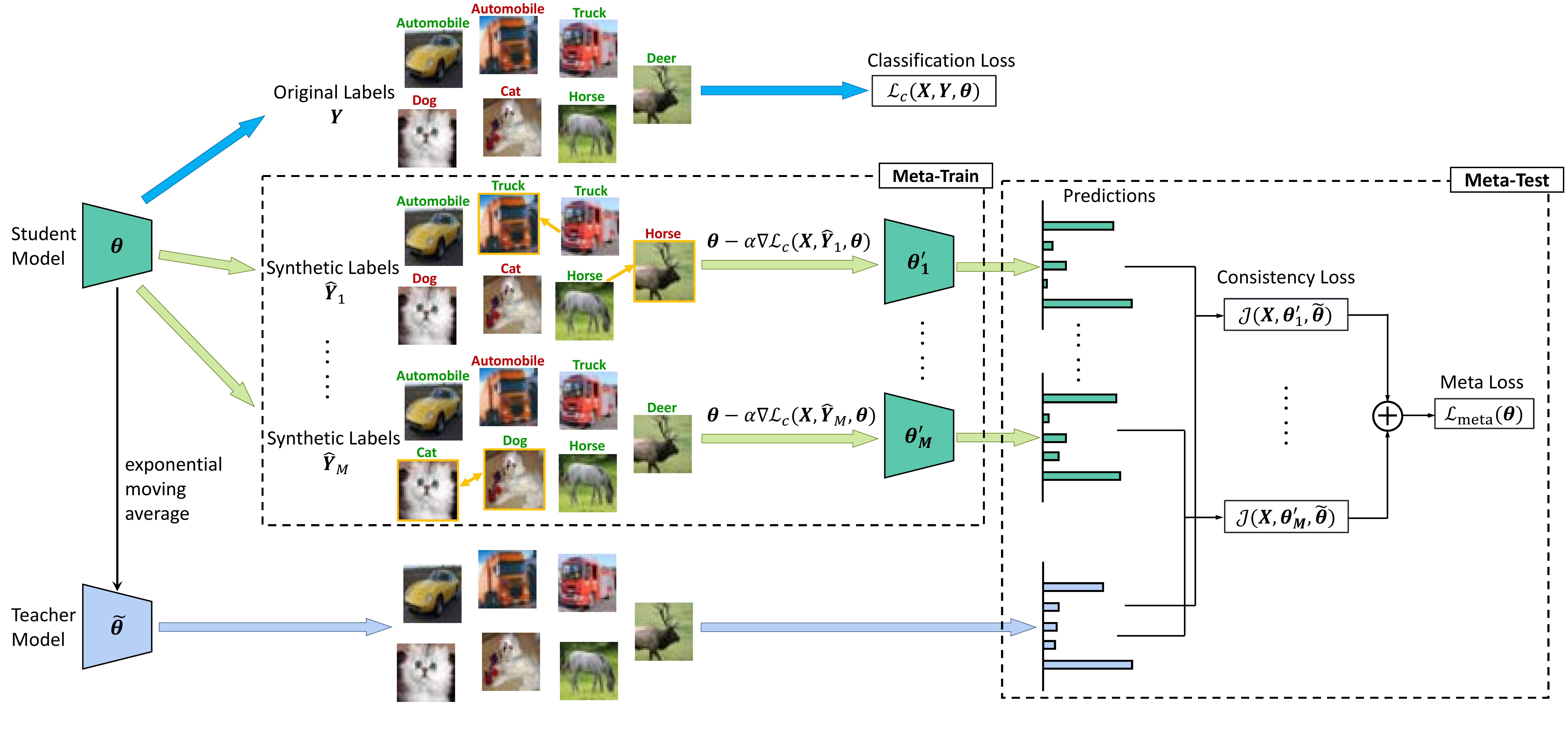}
  \caption
  {
  \small
	Illustration of the proposed meta-learning based noise-tolerant (MLNT) training. 
	For each mini-batch of training data,
	a meta loss is minimized before training on the conventional classification loss.
	We first generate multiple mini-batches of synthetic noisy labels with random neighbor label transfer (marked by \textcolor{orange}{orange} arrow).
	The random neighbor label transfer can preserve the underlying noise transition (\eg~{\footnotesize DEER $\rightarrow$ HORSE, CAT $\leftrightarrow$ DOG}),
	therefore generating synthetic label noise in a similar distribution as the original data.
	For each synthetic mini-batch,
	we update the parameters with gradient descent, and enforce the updated model to give consistent predictions with a \textit{teacher} model.
	The meta-objective is to minimize the consistency loss across all updated models \textit{w.r.t} $\Vec{\theta}$.
  } 
  \label{fig:framework}
 \end{figure*}  	
\subsection{Problem Statement}
We consider a classification problem with a training set
$\mathcal{D}=\{(\Vec{x}_1,\Vec{y}_1),...,(\Vec{x}_n,\Vec{y}_n)\}$,
where $\Vec{x}_i$ denotes the $i^{th}$ sample and $\Vec{y}_i \in \{0,1\}^c$ is a one-hot vector representing the corresponding noisy label over $c$ classes.
Let $f(\Vec{x}_i,\Mat{\theta})$ denotes the discriminative function of a neural network parameterized by $\Mat{\theta}$,
which maps an input to an output of the $c$-class softmax layer.
The conventional objective for supervised classification is to minimize an empirical risk, such as the cross entropy loss:
\begin{equation}
\label{eqn:ce}
	\mathcal{L}_c = -\frac{1}{n}\sum_{i=1}^{n} \Vec{y}_i \cdot \log(f(\Vec{x}_i,\Mat{\theta})),
\end{equation}
where $\cdot$ denotes dot product.

However, since $\Vec{y}_i$ contains noise,
the neural network can overfit and perform poorly on the test set.
We propose a meta-objective that encourages the network to learn noise-tolerant parameters.
The details are delineated next.

\subsection{Meta-Learning based Noise-Tolerant Training}
Our method can learn the parameters of a DNN model in such a way as to ``prepare'' the model for label noise.
The intuition behind our method is that when training with a gradient-based rule,
some network parameters are more tolerant to label noise than others.
How can we encourage the emergence of such noise-tolerant parameters?
We achieve this by introducing a meta-learning update before the conventional update for each mini-batch.
The meta-learning update simulates the process of training with label noise and makes the network less prone to over-fitting.
Specifically,
for each mini-batch of training data,
we generate a variety of synthetic noisy labels on the same images.
With each set of synthetic noisy labels,
we update the network parameters using one gradient update,
and enforce the updated network to give consistent predictions with a \textit{teacher} model unaffected by the synthetic noise.
As shown in \fig~\ref{fig:intuition},
the meta-learning update optimizes the model so that it can learn better with conventional gradient update on the original mini-batch.
In effect,
we aim to find model parameters that are less sensitive to label noise and can consistently learn the underlying knowledge from data despite label noise.
The proposed meta-learning update consists of two procedures: meta-train and meta-test.

\noindent\textbf{Meta-Train.}
Formally,
at each training step,
we consider a mini-batch of data $(\Mat{X},\Mat{Y})$ sampled from the training set,
where $\Mat{X} = \{\Vec{x}_1,...,\Vec{x}_k\}$ are $k$ samples,
and $\Mat{Y} = \{\Vec{y}_1,...,\Vec{y}_k\}$ are the corresponding noisy labels.
We want to generate multiple mini-batches of noisy labels $\{\hat{\Mat{Y}}_1,...,\hat{\Mat{Y}}_M\}$ with similar label noise distribution as $\Mat{Y}$.
We will describe the procedure for generating one set of noisy labels $\hat{\Mat{Y}}_m = \{\hat{\Vec{y}}_1^m,...,\hat{\Vec{y}}_k^m\}$.
First, 
we randomly select $\rho$ samples out of the mini-batch of $k$ samples.
For each selected sample $\Vec{x}_i$, 
we rank its neighbors within the mini-batch.
Then we randomly select a neighbor $\Vec{x}_j$ from its top 10 nearest neighbors (10 is experimentally determined),
and use the neighbor's label $\Vec{y}_j$ to replace the label for  $\Vec{x}_i$,
$\hat{\Vec{y}}_i^m = \Vec{y}_j$. 
Because we transfer labels among neighbors, the synthetic noisy labels are from a similar distribution as the original noisy labels.
We repeat the above procedure $M$ times to generate $M$ mini-batches of synthetic noisy labels.
Note that we compute nearest neighbors based on the Euclidean distance between feature representations (pre-softmax layer activations) generated by a DNN pre-trained on the entire noisy training set $\mathcal{D}$.

Let $\Vec{\theta}$ denote the current model's parameters,
for each synthetic mini-batch $(\Mat{X},\hat{\Mat{Y}}_m)$,
we update $\Vec{\theta}$ to $\Vec{\theta}_m'$ using one gradient descent step on the mini-batch.
\begin{equation}
\label{eqn:gd}
	\Vec{\theta}_m' = \Vec{\theta}-\alpha \nabla_{\theta} \mathcal{L}_c (\Mat{X},\hat{\Mat{Y}}_m,\Vec{\theta}),
\end{equation}
where $\mathcal{L}_c (\Mat{X},\hat{\Mat{Y}}_m,\Vec{\theta})$ is the cross entropy loss described in equation~\ref{eqn:ce},
and $\alpha$ is the step size.

\vspace{1ex}
\noindent\textbf{Meta-Test.}
Our meta-objective is to train $\Vec{\theta}$ such that each updated parameters $\Vec{\theta}_m'$ do not overfit to the specific noisy labels $\hat{\Mat{Y}}_m$.
We achieve this by enforcing each updated model to give consistent predictions with a \textit{teacher} model.
We consider the model parameterized by $\Vec{\theta}$ as the \textit{student} model,
and construct the \textit{teacher} model parameterized by $\tilde{\Vec{\theta}}$ following the self-ensembling~\cite{Tarvainen_NIPS_17} method.
The teacher model usually gives better predictions than the student model.
Its parameters are computed as the exponential moving average (EMA) of the student's parameters.
Specifically, at each training step, we update $\tilde{\Vec{\theta}}$ with:
\begin{equation}
\tilde{\Vec{\theta}} = \gamma \tilde{\Vec{\theta}}+(1-\gamma)\Vec{\theta},
\end{equation}
where $\gamma$ is a smoothing coefficient hyper-parameter.

Since the teacher model is unaffected by the synthetic label noise,
we enforce a consistency loss $\mathcal{J}(\Vec{\theta}_m')$ that encourages each updated model (with parameters $\Vec{\theta}_m'$) to give consistent predictions with the teacher model on the same input $\Vec{X}$.
We define $\mathcal{J}(\Vec{\theta}_m')$ as the Kullback-Leibler (KL)-divergence between the softmax predictions from the updated model $f(\Mat{X},\Vec{\theta}_m')$ and the softmax predictions from the teacher model $f(\Mat{X},\tilde{\Vec{\theta}})$.
We find that KL-divergence produces better results compared to the mean squared error used in~\cite{Tarvainen_NIPS_17}.
\begin{align}
\hspace{-1ex}\mathcal{J}(\Vec{\theta}_m')&=\frac{1}{k}\sum_{i=1}^{k} D_{KL}(f(\Vec{x}_i,\tilde{\Vec{\theta}}) || f(\Vec{x}_i,\Vec{\theta}_m')) \\
	&=\frac{1}{k}\sum_{i=1}^{k} \mathbb{E}(\log (f(\Vec{x}_i,\tilde{\Vec{\theta}})) - \log(f(\Vec{x}_i,\Vec{\theta}_m'))).
\end{align}

We want to minimize the consistency loss for all of the $M$ updated models with parameters $\{\Vec{\theta}_1',...,\Vec{\theta}_M'\}$.
Therefore, the meta loss is defined as the average of all consistency losses:
\begin{align}
\mathcal{L}_\mathrm{meta}(\Vec{\theta}) &=
\frac{1}{M}\sum_{m=1}^{M}\mathcal{J}(\Vec{\theta}_m') \\
\label{eqn:meta}
&=\frac{1}{M}\sum_{m=1}^{M} \mathcal{J}(\Vec{\theta}-\alpha \nabla_{\theta} \mathcal{L}_c (\Mat{X},\hat{\Mat{Y}}_m,\Vec{\theta})).
\end{align}

\begin{algorithm}[!t]
    \caption{Meta-Learning based Noise-Tolerant Training}
    \label{alg:meta}
    \begin{algorithmic}[1]
        \State Randomly initialize $\Vec{\theta}$
        \State Initialize teacher model $\tilde{\Vec{\theta}}=\Vec{\theta}$
        \While{not done}                	
        	\State Sample a mini-batch $(\Mat{X},\Mat{Y})$ of size $k$ from $\mathcal{D}$.
        	\For{$m=1:M$}
        		\parState{Generate synthetic noisy labels $\hat{\Mat{Y}}_m$ by random neighbor label transfer}
        		\parState{Compute updated parameters with gradient descent: $\Vec{\theta}_m' = \Vec{\theta}-\alpha \nabla_{\theta} \mathcal{L}_c (\Mat{X},\hat{\Mat{Y}}_m,\Vec{\theta})$  }   
        		\parState{Evaluate consistency loss with teacher:\\
        		 $\mathcal{J}(\Vec{\theta}_m')=\frac{1}{k}\sum_{i=1}^{k} D_{KL}(f(\Vec{x}_i,\tilde{\Vec{\theta}}) || f(\Vec{x}_i,\Vec{\theta}_m'))$}
        	\EndFor

        	\State Evaluate $\mathcal{L}_\mathrm{meta}(\Vec{\theta}) = \frac{1}{M}\sum_{m=1}^{M}\mathcal{J}(\Vec{\theta}_m')$
        	\State Meta-learning update $	\Vec{\theta} \leftarrow \Vec{\theta}-\eta\nabla \mathcal{L}_\mathrm{meta}(\Vec{\theta})$
        	\State Evaluate classification loss $\mathcal{L}_c (\Mat{X},\Mat{Y},\Vec{\theta})$    
        	\State Update $	\Vec{\theta} \leftarrow \Vec{\theta}-\beta\nabla \mathcal{L}_c (\Mat{X},\Mat{Y},\Vec{\theta})$        	    	
			\State Update teacher model: $\tilde{\Vec{\theta}} = \gamma \tilde{\Vec{\theta}}+(1-\gamma)\Vec{\theta}$        	
        \EndWhile        
    \end{algorithmic}
\end{algorithm}


Although $\mathcal{L}_\mathrm{meta}(\Vec{\theta})$ is computed using the updated model parameters $\Vec{\theta}_m'$, the optimization is performed over the student model parameters $\Vec{\theta}$.
We perform stochastic gradient descent (SGD) to minimize the meta loss.
The model parameters $\Vec{\theta}$ are updated as follows:
\begin{equation}
	\Vec{\theta} \leftarrow \Vec{\theta}-\eta\nabla \mathcal{L}_\mathrm{meta}(\Vec{\theta}),
\end{equation}
where $\eta$ is the meta-learning rate.

After the meta-learning update, 
we perform SGD to optimize the classification loss on the original mini-batch $(\Mat{X},\Mat{Y})$.
\begin{equation}
	\Vec{\theta} \leftarrow \Vec{\theta}-\beta\nabla\mathcal{L}_c (\Mat{X},\Mat{Y},\Vec{\theta}),
\end{equation}
where $\beta$ is the learning rate.
The full algorithm is outlined in Algorithm~\ref{alg:meta}.

Note that the meta-gradient $\nabla\mathcal{L}_\mathrm{meta}(\Vec{\theta})$ involves a gradient through a gradient, which requires calculating the second-order derivatives with respect to $\Vec{\theta}$.
In our experiments we use a first-order approximation by omitting the second-order derivatives,
which can significantly increase the computation speed.
The comparison in Section~\ref{sec:ablation} shows that this approximation performs almost as well as using second-order derivatives.
This provides another intuition to explain our method:
The first-order approximation considers the term $\alpha \nabla_{\theta} \mathcal{L}_c (\Mat{X},\hat{\Mat{Y}}_m,\Vec{\theta})$ in equation~\ref{eqn:meta} as a constant.
Therefore, we can consider the update $\Vec{\theta}-\alpha \nabla_{\theta} \mathcal{L}_c (\Mat{X},\hat{\Mat{Y}}_m,\Vec{\theta})$ as injecting data-dependent noise to the parameters,
and adding noise to the network during training has been shown by many studies to have a regularization effect~\cite{Laine_ICLR_2017,dropout}.

\subsection{Iterative Training}
We propose an iterative training scheme for two purposes:
(1) Remove samples with potentially wrong class labels from the classification loss $\mathcal{L}_c (\Mat{X},\Mat{Y},\Vec{\theta})$.
(2) Improve the predictions from the teacher model $f(\Vec{x}_i,\tilde{\Vec{\theta}})$ so that the consistency loss is more effective. 

First, we perform an initial training iteration following the method described in Algorithm~\ref{alg:meta},
and acquire a model with the best validation accuracy (usually the teacher).
We name that model as \textit{mentor} and use $\Mat{\theta}^*$ to denote its parameters.
In the second training iteration,
we repeat the steps in Algorithm~\ref{alg:meta} with two changes described as follows.

First, if the classification loss $\mathcal{L}_c (\Mat{X},\Mat{Y},\Vec{\theta})$ is applied to the entire training set $\mathcal{D}$,
samples with wrong ground-truth labels can corrupt training.
Therefore, we remove a sample from the classification loss if the mentor model assigns a low probability to the ground-truth class.
In effect,
the classification loss would now sample batches from a filtered training set $\mathcal{D}'$ which contains fewer corrupted samples. 
\begin{equation}
	\mathcal{D}'=\{(\Vec{x}_i,\Vec{y}_i)\in \mathcal{D}\mid\Vec{y}_i \cdot f(\Vec{x}_i,\Mat{\theta}^*)>\tau\},
\end{equation}
where $f(\Vec{x}_i,\Mat{\theta}^*)$ is the softmax prediction of the mentor model, and $\tau$ is a threshold to control the balance between the quality and quantity of $\mathcal{D}'$.

Second,
we improve the effectiveness of the consistency loss by merging the predictions from the mentor model and the teacher model to produce more reliable predictions.
The new consistency loss is:
\begin{align}
\mathcal{J}'(\Vec{\theta}_m')&=\frac{1}{k}\sum_{i=1}^{k} D_{KL}(\Vec{s}||f(\Vec{x}_i,\Vec{\theta}_m')),\\
\Vec{s} & =  \lambda f(\Vec{x}_i,\tilde{\Vec{\theta}}) +(1-\lambda)f(\Vec{x}_i,\Mat{\theta}^*)),
\end{align} 
where $\lambda$ is a weight to control the importance of the teacher model and the mentor model. 
It is ramped up from 0 to 0.5 as training proceeds.

We train for three iterations in our experiments.
The mentor model is the best model from the previous iteration.
We observe that further training iterations beyond three do not give noticeable performance improvement.

\section{Experiments}
\label{sec:experiments}
\begin{table*}[!t]
	\centering
	\caption
		{
		\small	
		Classification accuracy (\%) on CIFAR-10 test set for different methods trained with \textit{symmetric} label noise. 
		We report the mean and standard error across 5 runs.
		}
	\label{tbl:cifar-sn}
	\vspace{-1ex}
	\begin{small}
	\begin{tabular}{L{29ex}|C{12ex}|C{12ex}|C{12ex}|C{12ex}|C{12ex}|C{12ex}}
		\toprule	
		Method\hspace{25ex} & $r=0$ & $r=0.1$ &$r=0.3$ &$r=0.5$ &$r=0.7$ &$r=0.9$ \\
		\midrule
		Cross Entropy~\cite{Tanaka_CVPR_2018} & 93.5 &91.0 &88.4 & 85.0&78.4 &41.1\\
		Cross Entropy (reproduced) &91.84$\pm$0.05 &90.33$\pm$0.06 & 87.85$\pm$0.08&84.62$\pm$0.08 &78.06$\pm$0.16 &45.85$\pm$0.91 \\
		
		Joint Optimization~\cite{Tanaka_CVPR_2018} & 93.4 &92.7 &91.4 & 89.6&85.9 &58.0\\
		\midrule
		MLNT-student (1st iter.) & 93.18$\pm$0.07 & 92.16$\pm$0.05& 90.57$\pm$0.08&87.68$\pm$0.06&81.96$\pm$0.19 &55.45$\pm$1.11\\
		MLNT-teacher (1st iter.) & 93.21$\pm$0.07 & 92.43$\pm$0.05& 91.06$\pm$0.07&88.43$\pm$0.05 &83.27$\pm$0.22 &57.39$\pm$1.13\\
		\midrule
		MLNT-student (2nd iter.) & 93.24$\pm$0.09 & 92.63$\pm$0.07& 91.99$\pm$0.13& 89.71$\pm$0.07 &86.28$\pm$0.19 &58.21$\pm$1.09\\
		MLNT-teacher (2nd iter.) & 93.35$\pm$0.07 & 92.91$\pm$0.09& 91.89$\pm$0.06& 90.03$\pm$0.08 &86.24$\pm$0.18 &58.33$\pm$1.10\\
		\midrule	
		MLNT-student (3nd iter.) & 93.29$\pm$0.08 & 92.91$\pm$0.10& 92.02$\pm$0.09& 90.27$\pm$0.10 &86.95$\pm$0.17 &58.57$\pm$1.12\\
		MLNT-teacher (3nd iter.) & \textbf{93.52}$\pm$0.08 & \textbf{93.24}$\pm$0.08 &\textbf{92.50}$\pm$0.07 & \textbf{90.65}$\pm$0.09 &\textbf{87.11}$\pm$0.19 &\textbf{59.09}$\pm$1.12\\			
		\bottomrule	
  \end{tabular}
  \end{small}
\end{table*}

\subsection{Datasets}
We conduct experiments on two datasets,
namely CIFAR-10~\cite{cifar} and Clothing1M~\cite{Tong_CVPR_2015}. 
We follow the same experimental setting as previous studies~\cite{Giorgio_CVPR_2017,Tanaka_CVPR_2018,Vahdat_NIPS_2017} for fair comparison.

For CIFAR-10, we split 10\% of the training data for validation,
and artificially corrupt the rest of the training data with two types of label noise: symmetric and asymmetric.
The symmetric label noise is injected by using a random one-hot vector to replace the ground-truth label of a sample with a probability of $r$.
The asymmetric label noise is designed to mimic some of the structure of real mistakes for similar classes~\cite{Giorgio_CVPR_2017}:
{\small TRUCK $\rightarrow$ AUTOMOBILE,
BIRD $\rightarrow$ AIRPLANE,
DEER $\rightarrow$ HORSE,
CAT $\leftrightarrow$ DOG}.
Label transitions are parameterized by $r\in[0,1]$ such that true class and wrong class have probability of $1-r$ and $r$, respectively.

Clothing1M~\cite{Tong_CVPR_2015} consists of 1M images collected from online shopping websites, which are classified into 14 classes,~\eg t-shirt, sweater, jacket. 
The labels are generated using surrounding texts provided by sellers,
which contain real-world errors.
We use the $14k$ and $10k$ clean data for validation and test, respectively,
but we do not use the $50k$ clean training data.
\begin{table*}[!t]
	\centering
	\caption
		{
		\small	
		Classification accuracy (\%) on CIFAR-10 test set for different methods trained with \textit{asymmetric} label noise. 
		We report the mean and standard error across 5 runs.
		}
	\label{tbl:cifar-an}
	\vspace{-1ex}
	\begin{small}
	\begin{tabular}{L{32ex}|C{13ex}|C{13ex}|C{13ex}|C{13ex}|C{13ex}} 
		\toprule	
		Method & $r=0.1$ & $r=0.2$ &$r=0.3$ &$r=0.4$ &$r=0.5$  \\
		\midrule
		Cross Entropy~\cite{Tanaka_CVPR_2018}      & 91.8           & 90.8           & 90.0           & 87.1           & 77.3           \\
		Cross Entropy (reproduced)                 & 91.04$\pm$0.07 & 90.19$\pm$0.09 & 88.88$\pm$0.06 & 86.34$\pm$0.22 & 77.48$\pm$0.79 \\
		
		Forward~\cite{Giorgio_CVPR_2017}           & 92.4           & 91.4           & 91.0           & 90.3          & 83.8            \\
		CNN-CRF~\cite{Vahdat_NIPS_2017}            & 92.0           & 91.5           & 90.7           & 89.5          & 84.0 	          \\				
		Joint Optimization~\cite{Tanaka_CVPR_2018} & 93.2           & 92.7           & 92.4           & 91.5          & \textbf{84.6}   \\
		\midrule
		MLNT-student (1st iter.) & 92.89$\pm$0.11          & 91.84$\pm$0.10          & 90.55$\pm$0.09          & 88.70$\pm$0.13          & 79.95$\pm$0.71 \\
		MLNT-teacher (1st iter.) & 93.05$\pm$0.10          & 92.19$\pm$0.09          & 91.47$\pm$0.04          & 88.69$\pm$0.08          & 78.44$\pm$0.45 \\ 
		\midrule
		MLNT-student (2nd iter.) & 93.01$\pm$0.12          & 92.65$\pm$0.11          & 91.87$\pm$0.12          & 90.60$\pm$0.12          & 81.53$\pm$0.66 \\
		MLNT-teacher (2nd iter.) & 93.33$\pm$0.13          & 92.97$\pm$0.11          & 92.43$\pm$0.19          & 90.93$\pm$0.15          & 81.47$\pm$0.54 \\	
		\midrule	
		MLNT-student (3nd iter.) & 93.36$\pm$0.14          & 92.98$\pm$0.13          & 92.59$\pm$0.10          & 91.87$\pm$0.12          & 82.25$\pm$0.68 \\
		MLNT-teacher (3nd iter.) & \textbf{93.61}$\pm$0.10 & \textbf{93.25}$\pm$0.12 & \textbf{92.82}$\pm$0.18 & \textbf{92.30}$\pm$0.10 & 82.09$\pm$0.47 \\			
		\bottomrule
	\end{tabular}
	\end{small}
	\vspace{-1ex}
\end{table*}			

\subsection{Implementation}
For experiments on CIFAR-10, we follow the same experimental setting as~\cite{Tanaka_CVPR_2018} and use the network based on PreAct ResNet-32~\cite{He_ECCV_2016}.
By common practice~\cite{Tanaka_CVPR_2018},
we normalize the images,
and perform data augmentation by random horizontal flip and $32\times32$ random cropping after padding 4 pixels per side.
We use a batch size $k=128$, a step size $\alpha=0.2$, a learning rate $\beta=0.2$, and update $\Vec{\theta}$ using SGD with a momentum of 0.9 and a weight decay of $10^{-4}$.
For each training iteration, we divide the learning rate by 10 after 80 epochs, and train until 120 epochs.
For the initial iteration,
we ramp up $\eta$ (meta-learning rate) from 0 to 0.4 during the first 20 epochs,
and keep $\eta=0.4$ for the rest of the training.
In terms of the EMA decay $\gamma$, we use $\gamma=0.99$ for the first 20 epochs and $\gamma=0.999$ later on,
because the student improves quickly early in the training, and thus the teacher should have a shorter memory~\cite{Tarvainen_NIPS_17}.
In the ablation study (Section~\ref{sec:ablation}), we will show the effect of the three important hyper-parameters,
namely $M$, the number of synthetic mini-batches,
$\rho$, the number of samples with label replacement,
and the threshold $\tau$ for data filtering. 
The value for all hyper-parameters are determined via validation.

For experiments on Clothing1M,
we follow previous works~\cite{Giorgio_CVPR_2017,Tanaka_CVPR_2018} and use the ResNet-50~\cite{resnet} pre-trained on ImageNet.
For preprocessing,
we resize the image to $256\times256$,
crop the middle $224\times224$ as input,
and perform normalization.
We use a batch size $k=32$, a step size $\alpha=0.02$, a learning rate $\beta=0.0008$, and update $\Vec{\theta}$ using SGD with a momentum of 0.9 and a weight decay of $10^{-3}$.
We train for 3 epochs for each iteration.
During the first 2000 mini-batches in the initial iteration,
we ramp up $\eta$ from 0 to $0.0008$, and set $\gamma=0.99$.
For the rest of the training,
we use $\eta=0.0008$ and $\gamma=0.999$.
Other hyper-parameters are set as $M=10$,
$\rho=0.5k$, and $\tau=0.3$.

\subsection{Experiments on CIFAR-10}

We compare the proposed MLNT with multiple baseline methods using CIFAR-10 dataset with symmetric label noise (noise ratio $r=0,0.1,0.3,0.5,0.7,0.9$) 
and asymmetric label noise (noise ratio $r=0.1,0.2,0.3,0.4,0.5$). 
The baselines include:


\noindent(1)~\textbf{Cross Entropy}: conventional training without the meta-learning update. We report both the results from~\cite{Tanaka_CVPR_2018} and from our own implementation.

\noindent(2)~\textbf{Forward}~\cite{Giorgio_CVPR_2017}: forward loss correction based on the noise transition matrix.

\noindent(3)~\textbf{CNN-CRF}~\cite{Vahdat_NIPS_2017}: a CRF model is proposed to represent the relationship between noisy and clean labels. It requires a small set of clean labels during training.

\noindent(4)~\textbf{Joint Optimization}~\cite{Tanaka_CVPR_2018}: alternatively updates network parameters and corrects labels during training.

Both Forward~\cite{Giorgio_CVPR_2017} and CNN-CRF~\cite{Vahdat_NIPS_2017} require the ground-truth noise transition matrix, and Joint Optimization~\cite{Tanaka_CVPR_2018} requires the ground-truth class distribution among training data.
Our method does not require any prior knowledge on the data, thus is more general.
Note that all baselines use the same network architecture as our method.
We report the numbers published in~\cite{Tanaka_CVPR_2018}.  

Table~\ref{tbl:cifar-sn} and Table~\ref{tbl:cifar-an} show the results for symmetric and asymmetric label noise, respectively.
Our implementation of Cross Entropy has lower overall accuracy compared to~\cite{Tanaka_CVPR_2018}.
The reason could be the different programming frameworks used (we use PyTorch~\cite{pytorch}, whereas~\cite{Tanaka_CVPR_2018} used Chainer~\cite{chainer}).
For both types of noise, the proposed MLNT method with one training iteration significantly improves accuracy compared to Cross Entropy (reproduced),
and achieves comparable performance to state-of-the-art methods.
Iterative training further improves the performance.
MLNT-teacher after three iterations significantly outperforms previous methods.
An exception where MLNT does not outperform baselines is with 50\% asymmetric label noise.
This is because that asymmetric label noise is generated by exchanging {\small CAT} and {\small DOG} classes, and it is theoretically impossible to distinguish them without prior knowledge when the noise ratio is 50\%.

In Table~\ref{tbl:cifar-sn}, we also show the results with clean training data ($r=0$).
The proposed MLNT can achieve an improvement of $+1.68$ in accuracy compared to Cross Entropy,
which shows the regularization effect of the proposed meta-objective.

\begin{figure*}[!t]
 \centering
 	\hspace{4ex}
    \begin{subfigure}[t]{0.9\columnwidth}    	
    	\centerline{\includegraphics[width=\linewidth]{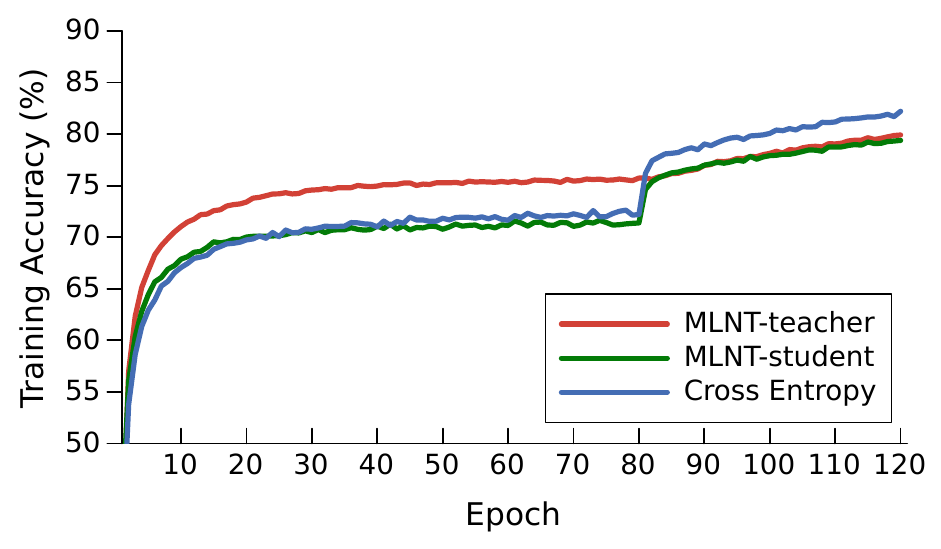}}
    	\vspace{-1ex}
    	  \caption{\small Training accuracy on noisy training data}
    	\label{fig:trend_train}
    \end{subfigure}
    \hfill
    \begin{subfigure}[t]{0.9\columnwidth}
    	\centerline{\includegraphics[width=\linewidth]{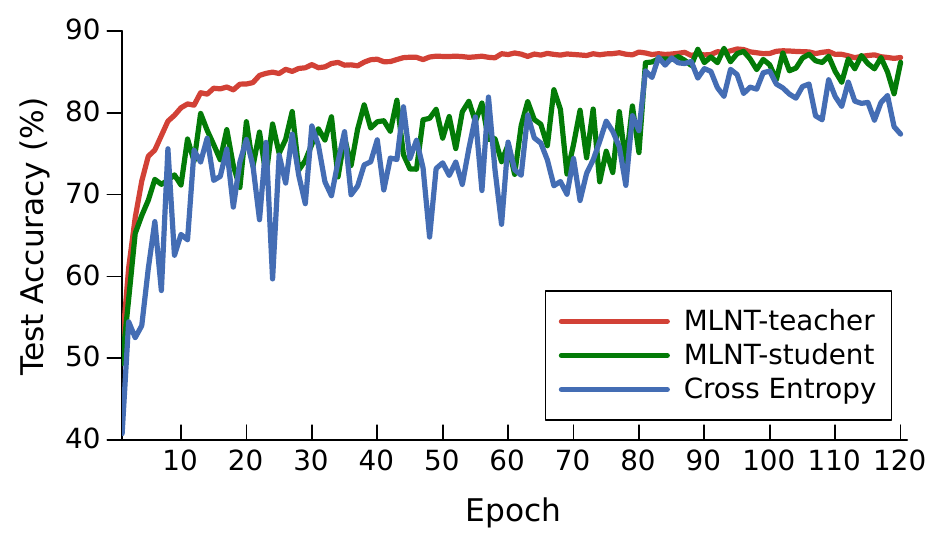}}
    	\vspace{-1ex}
    	 \caption{\small Test accuracy on clean test data}
    	\label{fig:trend_test}
    \end{subfigure}
	\hspace{5ex}
	\vspace{-1ex}
  \caption
	{
		\small
		Progressive performance comparison of the proposed MLNT and Cross Entropy as training proceeds.
	}
  \label{fig:trend}
  \vspace{-3ex}
\end{figure*}

\subsection{Ablation Study}
\label{sec:ablation}

\noindent\textbf{Progressive Comparison.}
Figure~\ref{fig:trend} plots the model's accuracy on noisy training data and its test accuracy on clean test data as training proceeds.
We show a representative training process using asymmetric label noise with $r=0.4$.
Accuracy is calculated every epoch,
and training accuracy is computed across all mini-batches within the epoch.
Comparing the proposed MLNT methods (1st iter.) with Cross Entropy,
MLNT learns more quickly during the beginning of training,
as shown by the higher test accuracy.
Cross Entropy has the most unstable training process,
as shown by its fluctuating test accuracy curve. 
MLNT-student is more stable because of the regularization from the meta-objective,
whereas MLNT-teacher is extremely stable because its parameters change smoothly during training.
At the 80th epoch,
the learning rate is divided by 10,
which causes a drastic increase in both training and test accuracy for MLNT-student and Cross Entropy.
After the 80th epoch,
the model begins to overfit because of the small learning rate.
However,
the proposed MLNT-student suffers less overfitting compared to Cross Entropy,
as shown by its lower training accuracy and higher test accuracy.

\begin{figure}[!t]
	\centering
  \includegraphics[trim={0 6 0 0},clip,width=0.9\columnwidth]{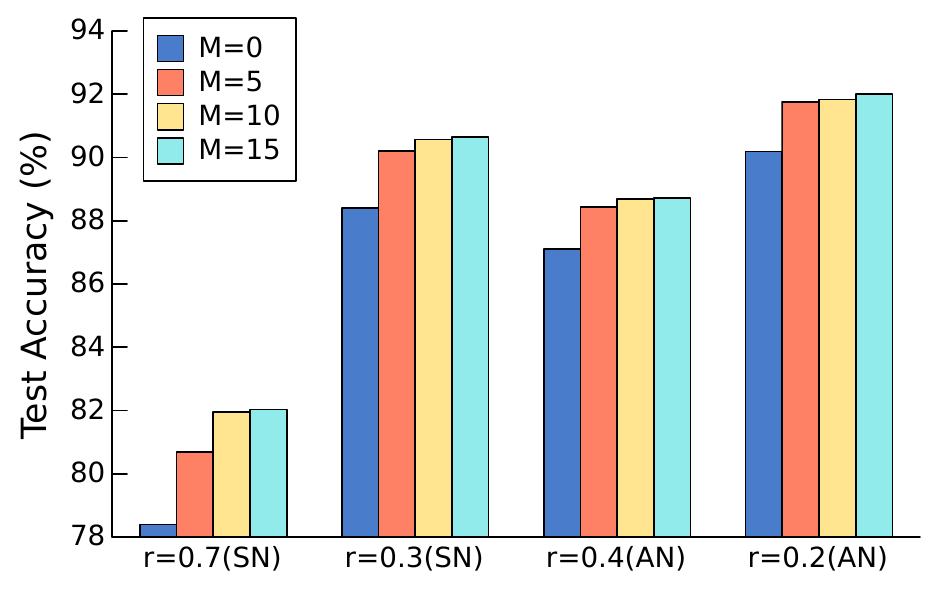}
  \caption
	{
		\small
		Performance of MLNT-student (1st iter.) on CIFAR-10 trained with different number of synthetic mini-batches $M$.
	}
\label{fig:M}
\vspace{-1ex}
\end{figure}

\noindent\textbf{Hyper-parameters.}
We conduct ablation study to examine the effect of three hyper-parameters: $M,\rho,\tau$.
$M$ is the number of mini-batches $\{(\Mat{X},\hat{\Mat{Y}}_m)\}_{m=1}^M$ with synthetic noisy labels that we generate for each mini-batch $(\Mat{X},\Mat{Y})$ from the original training data.
Intuitively, with larger $M$, the model is exposed to a wider variety of label noise, and thus can learn to be more noise-tolerant.
In Figure~\ref{fig:M} we show the test accuracy on CIFAR-10 for MLNT-student (1st iter.) with $M=0,5,10,15$ ($M=0$ is the same as Cross Entropy) trained using labels with symmetric noise (SN) and asymmetric noise (AN) of different ratio.
The result shows that the accuracy indeed increases as $M$ increases. 
The increase is most significant when $M$ changes from 0 to 5,
and is marginal when $M$ changes from 10 to 15.
Therefore, the experiments in this paper are conducted using $M=10$,
as a trade-off between the training speed and the model's performance.

\begin{figure}[!t]
 	\centering
  \includegraphics[trim={0 6 0 0},clip,width=0.9\columnwidth]{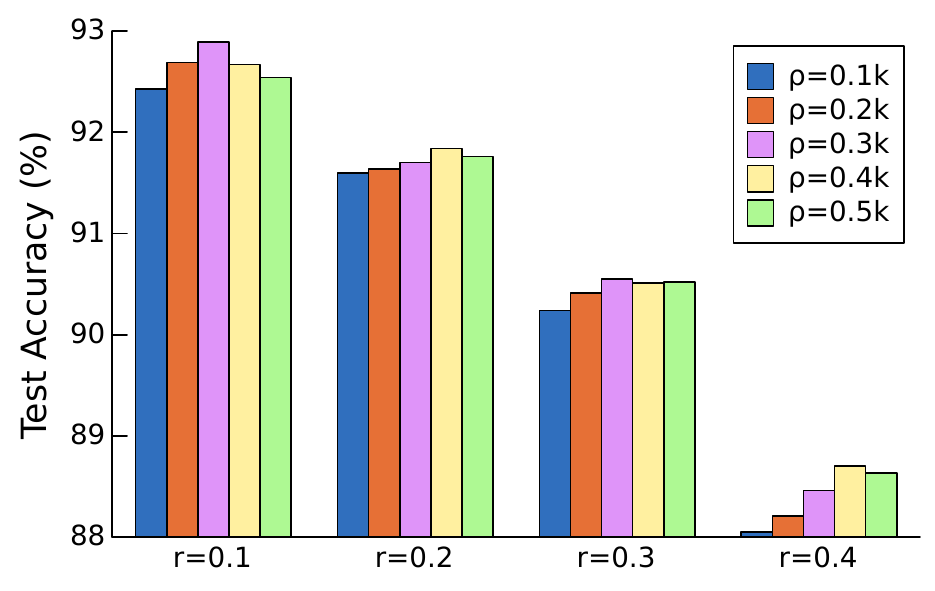}
  \caption
	{
		\small
		Performance of MLNT-student (1st iter.) on CIFAR-10 trained with asymmetric label noise using different $\rho$.
	}
\label{fig:rho}
\vspace{-1ex}
\end{figure}

\begin{figure}[!t]
  \centering
  \includegraphics[trim={0 6 0 0},clip,width=0.9\columnwidth]{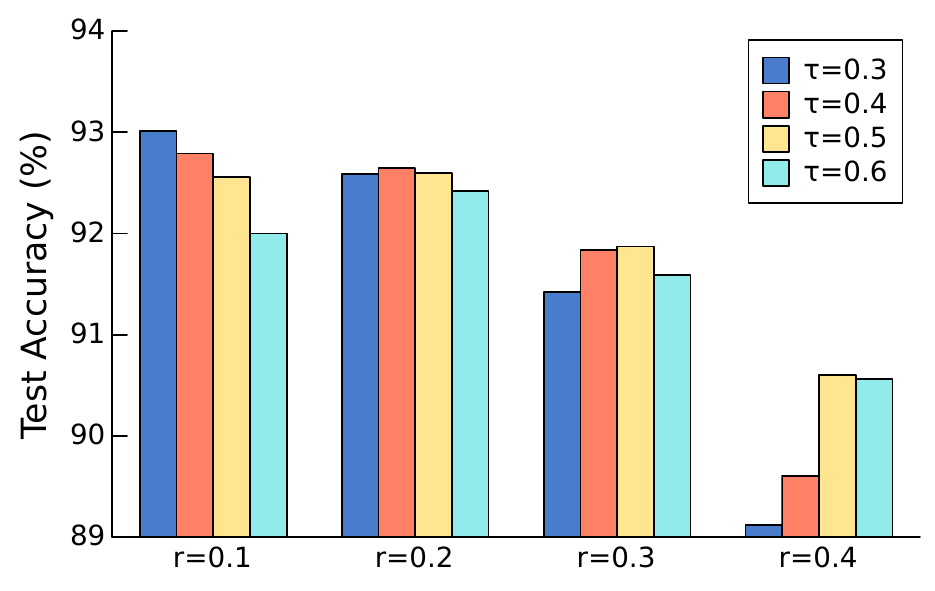}
  \caption
	{
		\small
		Performance of MLNT-student (2nd iter.) on CIFAR-10 trained with asymmetric label noise using different $\tau$.
	}
\label{fig:tau}
\vspace{-1ex}
\end{figure}

$\rho$ is the number of samples whose labels are changed in each synthetic mini-batch $\hat{\Mat{Y}}_m$ of size $k$.
We experiment with $\rho=0.1k,0.2k,0.3k,0.4k,0.5k$, which correspond to $13,26
,39,51,64$ samples with a batch size of 128.
Figure~\ref{fig:rho} shows the performance of MLNT-student (1st iter.) using different $\rho$ trained on CIFAR-10 with different ratio of asymmetric label noise.
The performance is insensitive to the value of $\rho$.
For different noise ratio, the optimal $\rho$ generally falls into the range of $[0.3k,0.5k]$.

\begin{figure}[!t]
	\centering
	\includegraphics[trim={0 9 0 0},clip,width=1\columnwidth]{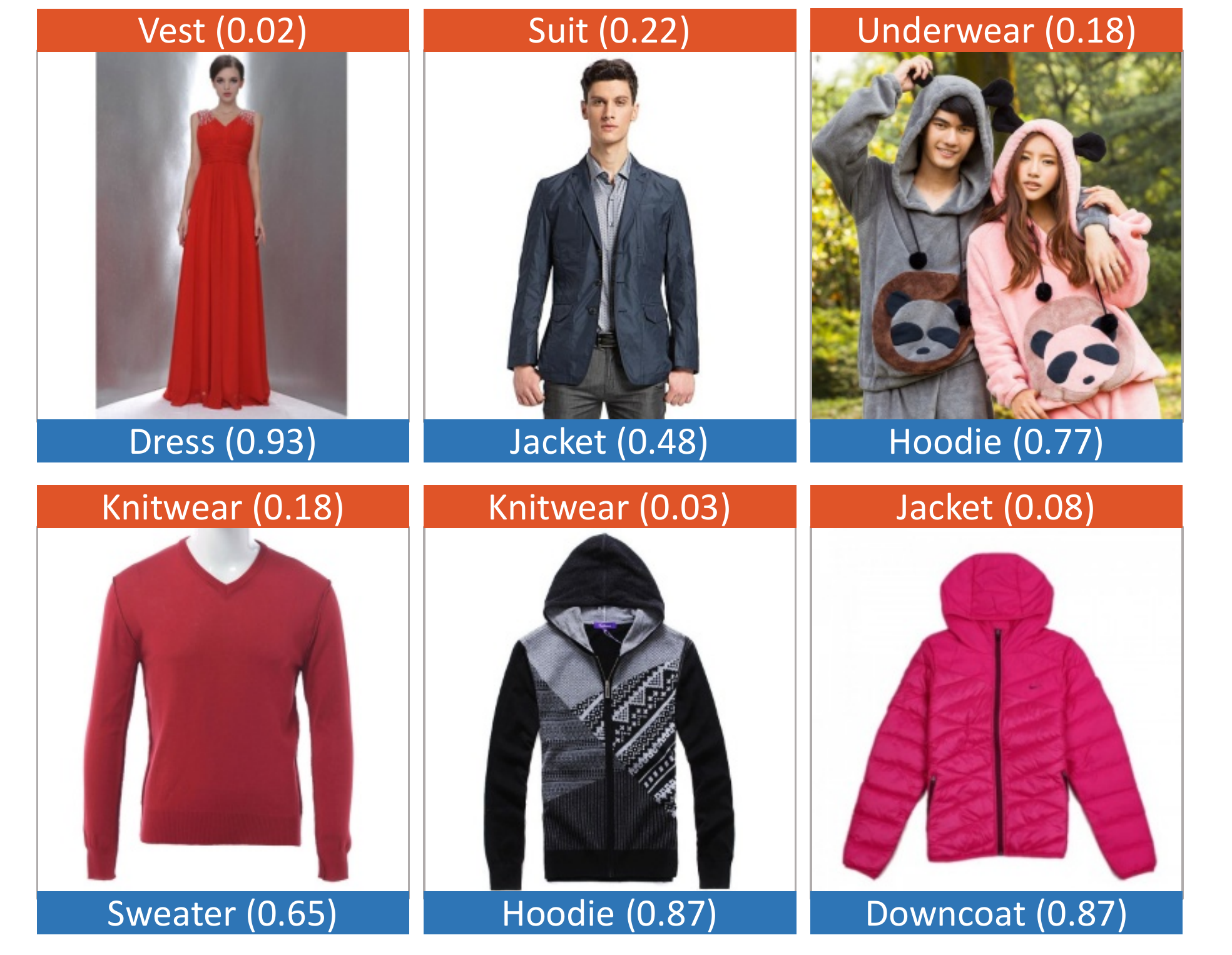}
	\caption
		{
			\small
		Example images from Clothing1M that are filtered out by the mentor model. We show the ground-truth label (red) and the label predicted by the mentor (blue) with their corresponding probability scores.
		}
	
	\label{fig:clothing}
\end{figure}

$\tau$ is the threshold to determine which samples are filtered out by the mentor model during the 2nd and 3nd training iteration.
It controls the balance between the quality and quantity of the data that is used by the classification loss.
In Figure~\ref{fig:tau} we show the performance of MLNT-student (2nd iter.) trained using different value of $\tau$. 
As the noise ratio $r$ increases, the optimal value of $\tau$ also increases to filter out more samples. 
Figure~\ref{fig:clothing} shows some example images from Clothing1M dataset that are filtered out and their corresponding probability scores given by the mentor model.


\noindent\textbf{Full optimization.}
We have been using a first-order approximation to optimize $\mathcal{L}_\mathrm{meta}$ for faster computation speed.
Here we conduct experiments using full optimization by including the second-order derivative with respect to $\Vec{\theta}$.
Table~\ref{tbl:second} shows the comparison on four representative sets of experiments with different label noise.
We show the test accuracy (averaged across 5 runs) of MLNT-student (1st iter.) trained with full optimization and first-order approximation.
The result shows that the performance from first-order approximation is nearly the same as that obtained with full second derivatives.
This suggests that the improvement of MLNT mostly comes from the gradients of the meta loss at the updated parameter values,
rather than the second-order gradients.

\subsection{Experiments on Clothing1M}

We demonstrate the efficacy of the proposed method on real-world noisy labels using the Clothing1M dataset.
The results are shown in Table~\ref{tbl:clothing}.
We show the accuracy for baselines \#1 and \#3 reported in~\cite{Tanaka_CVPR_2018},
and the accuracy for \#2 reported in~\cite{Giorgio_CVPR_2017}.
The proposed MLNT method with one training iteration achieves better performance compared to state-of-the-art methods.
After three training iterations, MLNT achieves a significant improvement in accuracy of $+4.19$ over Cross Entropy,
and an improvement of $+1.31$ over the best baseline method \#3. 

\begin{table}[!t]
	\centering
	\caption
		{
	  \small
		Test accuracy (\%) on CIFAR-10 for MNLT-student (1st iter.) with full optimization of the meta-loss and its first-order approximation.
		}
	\label{tbl:second}
	\vspace{-1ex}
	\resizebox{\columnwidth}{!}{
	\begin{tabular}{l|c|c|c|c} 
		\toprule	
		\multirow{2}{1.8cm}{Optimization} & \multicolumn{2}{c|}{SN}& \multicolumn{2}{c}{AN} \\
		\cmidrule{2-5}
		& $r=0.3$ & $r=0.7$ &$r=0.2$&$r=0.4$\\
		\midrule
		First-order approx.& 90.57 & 81.96 & 91.84 & 88.70\\
		Full& 90.74 & 82.05 & 91.89 & 88.91\\
		\bottomrule
	\end{tabular}
}
\end{table}			

\begin{table}[!t]
	\centering
	\caption
		{
			\small	
			Classification accuracy (\%) of different methods on the Clothing1M test set.
		}
	\label{tbl:clothing}
	\vspace{-1ex}
	\begin{small}
	\begin{tabular}{L{30ex}|C{13ex}} 
		\toprule	
		Method                                          & Accuracy \\
		\midrule
		\#1 Cross Entropy~\cite{Tanaka_CVPR_2018} 			& 69.15    \\
		Cross Entropy (reproduced) 											& 69.28    \\
		\#2 Forward~\cite{Giorgio_CVPR_2017} 						& 69.84    \\
		\#3 Joint Optimization~\cite{Tanaka_CVPR_2018} 	& 72.16    \\
		\midrule
		MLNT-student (1st iter.) & 72.34\\
		MLNT-teacher (1st iter.) & 72.08\\
		\midrule
		MLNT-student (2nd iter.) & 73.13\\
		MLNT-teacher (2nd iter.) & 73.10\\
		\midrule	
		MLNT-student (3nd iter.) & 73.44\\
		MLNT-teacher (3nd iter.) & \textbf{73.47}\\			
		\bottomrule
	\end{tabular}
	\end{small}
\end{table}			

\section{Conclusion}
\label{sec:conclusion}

In this paper, we propose a meta-learning method to learn from noisy labeled data,
where a meta-learning update is performed prior to conventional gradient update.
The proposed meta-objective aims to find noise-tolerant model parameters that are less prone to overfitting.
In the meta-train step,
we generate multiple mini-batches with synthetic noisy labels,
and use them to update the parameters.
In the meta-test step,
we apply a consistency loss between each updated model and a teacher model,
and train the original parameters to minimize the total consistency loss.
In addition,
we propose an iterative training scheme,
where the model from previous iteration is used to clean data and refine predictions.
We evaluate the proposed method on two datasets.
The results validate the advantageous performance of our method compared to state-of-the-art methods.
For future work,
we plan to explore using the proposed model-agnostic method to other domains with different model architectures,
such as learning Recurrent Neural Networks for machine translation with corrupted ground-truth sentences.

\section*{Acknowledgment}
\vspace{-0.5ex}
This research is supported by the National Research Foundation, Prime Minister's Office, Singapore under its Strategic Capability Research Centres Funding Initiative.

{\small
\balance
\bibliographystyle{ieee}
\bibliography{bib}
}

\end{document}